\documentclass[runningheads]{llncs}
\usepackage{graphicx}

\usepackage{epsfig}
\usepackage{graphicx}
\usepackage{amsmath}
\usepackage{amssymb}

\usepackage{bigstrut}
\usepackage{multirow}
\usepackage{times}
\usepackage{epsfig}
\usepackage{graphicx}
\usepackage{amsmath}
\usepackage{amssymb}
\usepackage{bigstrut}
\usepackage{multirow}
\usepackage{graphicx}
\usepackage{graphicx}
\usepackage{caption}
\usepackage{mathtools}
\usepackage{wrapfig}
\usepackage{float}
\usepackage{amsmath}
\usepackage{dblfloatfix}
\usepackage{mathrsfs}
\usepackage{url}
\usepackage{balance}
\usepackage{subfigure}
\usepackage{color}
\usepackage{epsfig}
\usepackage{graphicx}
\usepackage{hyperref}
\usepackage{graphicx}
\usepackage{graphics}
\usepackage{amsmath}
\usepackage{algorithm,algpseudocode}
\usepackage{amsmath}
\usepackage{amsfonts}
\usepackage{amssymb}

\makeatletter
\newcommand{\printfnsymbol}[1]{%
  \textsuperscript{\@fnsymbol{#1}}%
}
\makeatother

\begin{document}
\title{A Survey on Unknown Presentation Attack Detection for Fingerprint}
%
%

\author{Jag Mohan Singh \quad Ahmed Madhun \quad Guoqiang Li  \quad Raghavendra Ramachandra}
\authorrunning{Singh et al.}
\institute{Norwegian Biometrics Laboratory, Norwegian University of Science and Technology (NTNU), Norway \\
\email{jag.m.singh@ntnu.no},\email{ahmed.madhun@ntnu.no}, \email{guoqiang.li@ntnu.no}, \email{raghavendra.ramachandra@ntnu.no}
}

\maketitle              
\begin{abstract}
Fingerprint recognition systems are widely deployed in various real-life applications as they have achieved high accuracy. The widely used applications include border control, automated teller machine (ATM), and attendance monitoring systems. However, these critical systems are prone to spoofing attacks (a.k.a presentation attacks (PA)). PA for fingerprint can be performed by presenting gummy fingers made from different materials such as silicone, gelatine, play-doh, ecoflex, 2D printed paper, 3D printed material, or latex.
Biometrics Researchers have developed Presentation Attack Detection (PAD) methods as a countermeasure to PA. PAD is usually done by training a machine learning classifier for known attacks for a given dataset, and they achieve high accuracy in this task. However, generalizing to unknown attacks is an essential problem from applicability to real-world systems, mainly because attacks cannot be exhaustively listed in advance. In this survey paper, we present a comprehensive survey on existing PAD  algorithms for fingerprint recognition systems, specifically from the standpoint of detecting unknown PAD. We categorize PAD algorithms, point out their advantages/disadvantages, and future directions for this area.
\end{abstract}

\section{Introduction}

\begin{figure}[t!]
\centering
\includegraphics[width=1.0\linewidth]{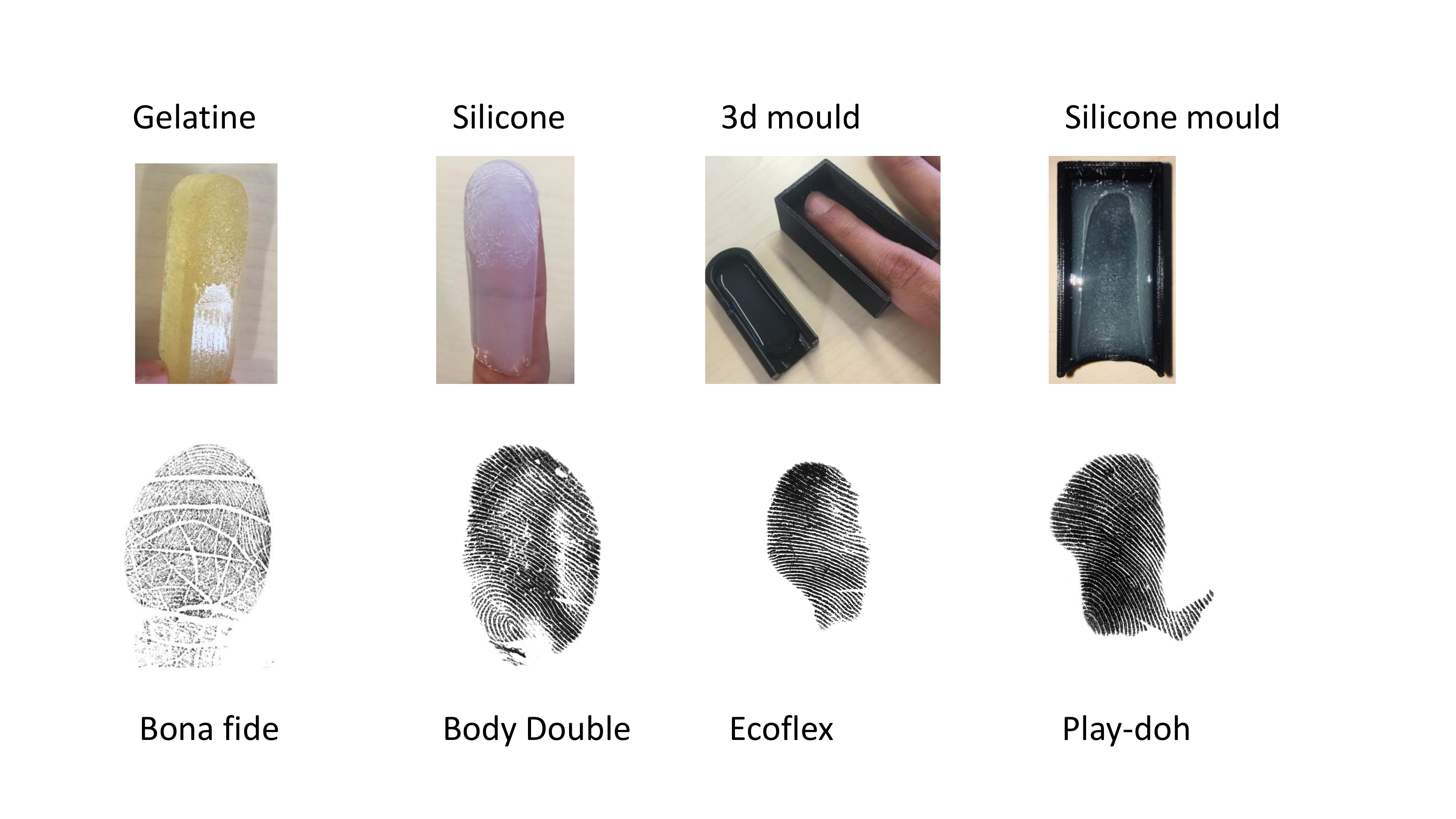} 
 \caption{Different attacks on the fingerprint recognition systems shown as photographs~\cite{9}, and as fingerprints~\cite{LivDet2015}).}
 \label{fig:fingersandattacks}
\end{figure}
Biometrics based authentication systems provide more security than traditional information security-based systems based on passwords/Personal Identification Number (PINs), and keys/cards~\cite{1}. The primary limitations with traditional information security methods are that they lack good user experience, using the same security measure with multiple applications, and forgetting/losing the password/PINs~\cite{2}. Especially for keys/cards, they can be duplicated apart from the previously mentioned limitations. Since biometric systems are based on human characteristics such as the face, fingerprint, or iris, which are unique for every individual, they have a definite advantage over information security-based systems. Due to these advantages, biometric systems are widely deployed in smartphones, border control (both in automated, and attended scenarios), and national identity cards. However, biometric systems are vulnerable to Presentation Attacks (PA)~\cite{3}, due to which some crimes have been reported in the media, where the biometric systems were spoofed~\cite{4,5,6}. An attacker can perform the attack on the biometric system by presenting a biometric artefact or a Presentation Attack Instruments (PAIs) ~\cite{RaghuSurvey}. PA can be performed in different biometric modalities, including the face, fingerprint, and iris. Since fingerprint recognition systems are widely deployed in critical security systems, it is essential to develop fingerprint PAD.


PAIs for fingerprint can either be an artificial object such as a gummy finger (made from play-doh, silicone, or gelatine) or a 2D/3D printed photo. In terms of implementation, PAD systems can be either a hardware-based or a software-based, whose main task is to distinguish between a real (bona fide) user or a malicious (imposter) attacker~\cite{8}. 
A summary of existing fingerprint PAD methods can be found in  Marcel et al.~\cite{10}, Marasco et al.~\cite{RossSurveyFPAD}, Galbally et al.~\cite{Galbally2019}, and Sousedik et al.~\cite{16}. In the current scenario, the majority of the existing PAD methods consist of training a classifier to accurately model the characteristics of the PAI. However, such an approach suffers from the problem of generalization to detect unknown attacks~\cite{10}. Thus, developing a reliable PAD technique for unknown attacks is a significant problem that can also be posed as anomaly (outlier) detection. Fingerprint recognition systems have been widely deployed, as mentioned earlier, and are prone to PA. Since the attacks cannot be listed in advance, detecting unknown attacks for the fingerprint is critical.
Our survey on fingerprint Presentation Attack Detection (FPAD) presents the following:
\begin{itemize}
    \item Comprehensive survey of existing methods for FPAD  for unknown attacks.
    \item Categorization of existing methods for the FPAD of unknown attacks.
    \item Discussion on advantages/disadvantages of existing methods for FPAD, especially for unknown attacks.
    \item Concluding remarks with future directions for the area of FPAD.
\end{itemize}

In the rest of the paper, a comparison between traditional PAD done in a supervised manner, and anomaly detection based FPAD in Section~\ref{sec:2}, which is followed by Section~\ref{sec:3} summarizing related work in FPAD  which includes their categorization, advantages, and disadvantages in terms of generalization, and finally we present conclusions \& future directions for FPAD in Section~\ref{sec:4}.

\section{Traditional PAD \& Anomaly Detection based PAD}
\label{sec:2}
\begin{figure}[htbp!]
\centering
\includegraphics[width=1.0\linewidth]{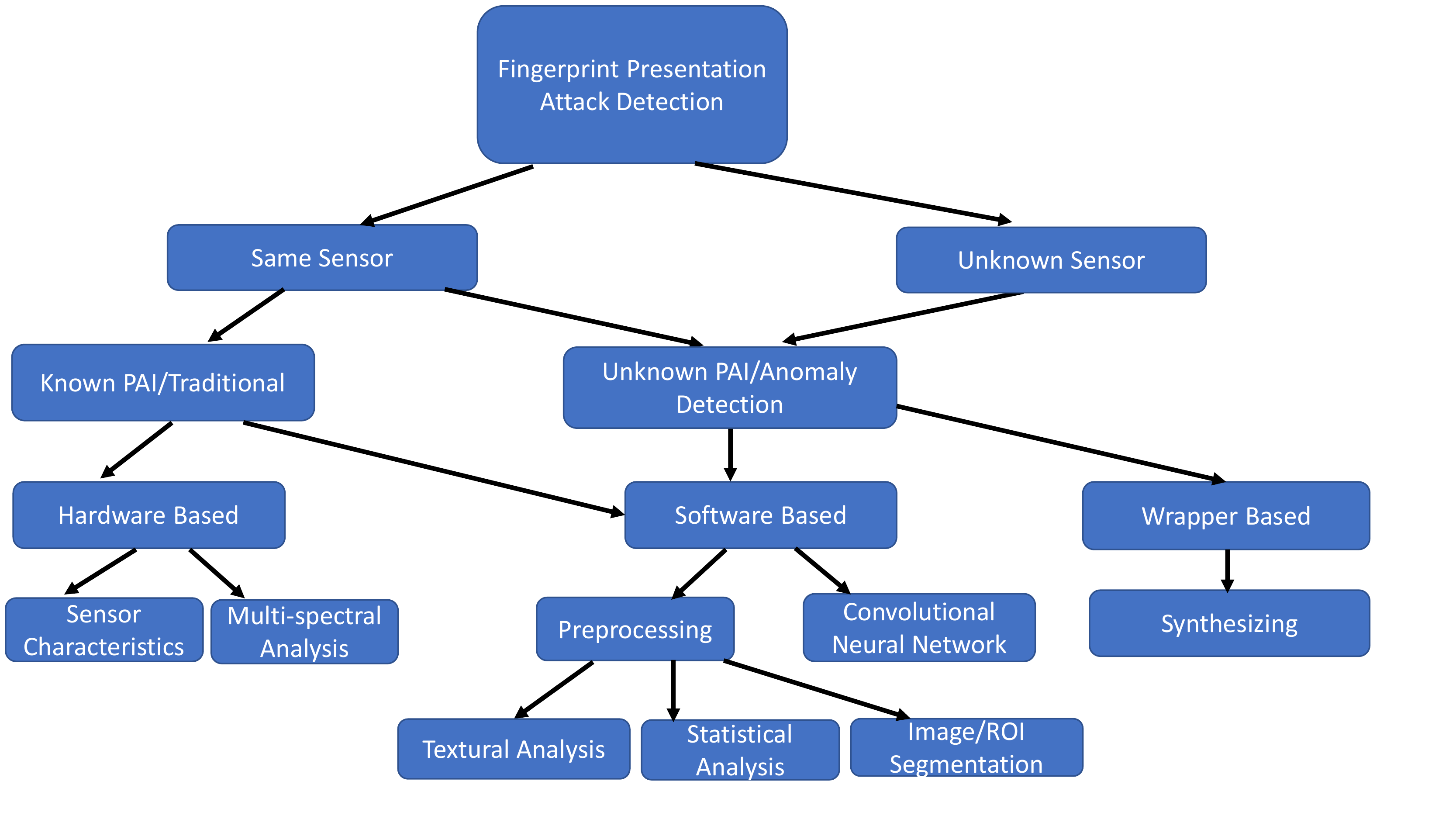} 
 \caption{Illustration of taxonomy for fingerprint presentation attack detection.}
 \label{fig:taxonomyfigure}
\end{figure}
\begin{table}[t!]
\centering
\resizebox{1\linewidth}{!}
{\begin{tabular}{|p{2.0cm}|l p{4.3cm}|l p{4.3cm}|}
\hline
 & & {\bf{Traditional PAD}} & & {\bf{Anomaly detection based PAD}} \\
\hline
{\bf{Characteristics}} &-& Information about PAI is gathered and known in advance. & - & Establish profiles of normality features which are extracted from regular data.  \\
 &-& Look for PAIs' features each time a presentation occurs. & - & Compares the normality features of each new presentation against the established profiles. \\
&-& Alerts for PA if any PAI is found to be in the new presentation.   & - & Alerts for PA if a deviation from normality is detected in the new presentation based on threshold.  \\

\hline
{\bf{Advantages}} &-& Possibility to detect known PAs. &-& Possibility to detect known and unknown PAs. \\
 &-& There is a possibility of using existing knowledge to recognize new forms of old PAs.  &-& Does not care about the used PAI during the attack. \\

\hline
{\bf{Drawbacks}} &-& For each novel PA, PAD methods should be updated and tested with the new PAI. &-& Hard to define a profile of normality features for each bona fide presentation. \\
 &-&
 As the number of PAs increases, and correspondingly PAIs increase, the complexity of PAD increases.
 
 &-&  Higher false-positive for PAs depending on accessibility or usability.  \\
 &-& Hard to detect previously unseen PAs. &-& Hard to set the optimal threshold value for PAD.\\
 &-& Simple changes to PAI in a known PA can be enough to miss the detection of the PA. &-& The size of normality feature can be very large, which leads to a high false-positive rate.\\
 &-& A leak of PAIs' list that a system maintains could help attackers bypass the system's PAD method. &&  \\
\hline
\end{tabular}}
\caption{Characteristics, advantages and disadvantages of Anomaly detection based PAD as compared to Traditional PAD for biometrics.\label{tab1}}
\end{table}
In this section, we present a comparison between traditional PAD (a form of supervised classification) and anomaly detection (supervised/unsupervised classification) based PAD, as shown in Figure~\ref{fig:taxonomyfigure}. Since we are interested in unknown attack detection for fingerprint, this can be achieved by Anomaly detection \cite{11}. We now briefly review Anomaly detection in the following subsection:
\subsection{Anomaly Detection}
Anomaly Detection refers to the determination of irregularity in a dataset. The dataset contains a set of records (aka., instances, objects, or entities), where each record includes a set of attributes (aka., characteristics or features), as pointed out by Chandola et al.~\cite{11}. In general, an anomaly detection method is provided with a record/set of records as an input, where no information about either anomalies or regular classes is known to the detection method in advance~\cite{14}.
The three modes of anomaly detection methods, according to Chandola et al.~\cite{11} are as follows:
\begin{itemize}
    \item \textit{\bf{Supervised anomaly detection}}: \\
    Anomaly methods that are based on a predictive model which is trained by a labeled dataset of two classes (i.e., normal and anomaly records). Any unseen record is then compared against the model to determine whether it is a normal or an anomaly record. This can be achieved by using publicly labeled fingerprint datasets for training. This form of anomaly detection is used in traditional PAD and in unknown attack detection where the sensor is known in advance for FPAD. 
    \item \textit{\bf{Semi-supervised anomaly detection}}:\\
    Anomaly methods that are based on a single classifier trained using only normal behavior records from a dataset, as only those are labeled. This form of anomaly detection is used for unknown attack detection both for the known sensor, \& unknown sensor for FPAD.
    \item \textit{\bf{Unsupervised anomaly detection}}:\\
    Anomaly methods do not require training data, but no records are labeled in the dataset if training is applied. This method is based on the assumption that regular records are far more frequent than anomalies in both training and testing datasets, and can lead to high false reject rate if this assumption is violated. This form of anomaly detection is used for unknown attack detection for the known sensor, and the unknown sensor for FPAD.
\end{itemize}
Table~\ref{tab1} shows a description  of traditional  and anomaly detection based PAD. Theoretically, the main advantage of anomaly-based PAD methods over traditional methods is capturing both the known and the unknown PAs. In contrast, the traditional PAD methods can detect known PAs, and maybe new forms of these attacks. For instance, if a traditional PAD method is trained only to detect gummy fingers of play-doh, it may not detect gummy fingers of other materials like silicone or gelatine. This requires the traditional PAD methods to make a long list of PAIs gathered from known PAs, and the methods should be updated and re-trained each time a new unknown PA is revealed. Consequently, the list of PAIs and known PAs can become long and hard to maintain. Moreover, if an attacker gets access to the list of PAIs used to train a biometric system, the attacker will be able to conduct a PA using a novel PAI that is not known to the systems.

Even if anomaly PAD methods solve several drawbacks in traditional PAD methods, they come with high risks, implementation difficulties, and critical disadvantages. In general, it is difficult to define and extract all features of bona fide presentations (i.e., normality features), because these features can have a broad scope and thus become hard to use for an implementation of a PAD method. Moreover, the threshold used to distinguish between PAs and bona fide presentations is affected with accessibility or usability issues between the subject, and the capture device, which makes it hard to define. Thus, the size of the normality features will be large, and it may require prioritizing some features over others during the feature selection. Nevertheless, size reduction methods can be used to reduce the number of features normality. However, this will lead to more false-positive alarms as the normality features are not precise enough to distinguish between all the cases of PAs and bona fide presentations.

\section{Known \& Unknown Presentation Attack Detection for fingerprints}
\label{sec:3}
\begin{table}[t!]
    \centering
    \resizebox{1\linewidth}{!}
    {\begin{tabular}{|l|l|l|l|l|l|l|l|}
        \hline
        \textbf{Ref.} & \textbf{S/H/W} & \textbf{Dataset} & \textbf{Pre-processing} & \textbf{Post-processing} & \textbf{\# unknown PAs} & \textbf{A. D. methods} & \textbf{A. D. mode}\\
        \hline
        \cite{fPA10} & S & LivDet 2009 & - & - & 1 & - & S \\ \hline
        
        \cite{fPA12} & S & LivDet 2011 & GLCM, HOG, BSIF & - & 2 & SVM, & S \\ 
        &  &  & LPQ, LBP, BGP &  &  & Rule-based &  \\ \hline
        
        \cite{fPA13} & S & LivDet 2011 & LBP & Score fusion & 4 & SVM  & S \\ \hline
        
        \cite{fPA11} & S & LivDet 2011 & BSIF, LBP, LPQ & - & 3 & SVM & S \\ \hline
        
        \cite{fPA3} & S & LivDet 2011, & Image segmentation & - & 4 & CNN & S \\ 
        &  & LivDet 2013, & (part of CNN) &  &  &  &  \\ 
        &  & LivDet 2015 &  &  &  &  &  \\ \hline
        
        \cite{fPA5} & S \& H & Own dataset & ROI segmentation & - & 6 & Pre-trained CNN & S \\ \hline
        
        \cite{fPA6} & S \& H & Own dataset & ROI segmentation & Score fusion & 3 & SVM & S \\ \hline
        
        \cite{fPA7} & S \& H & Own dataset & ROI segmentation & Score fusion & 5 & SVM, & S \\
        &  &  &  &  &  & CNN, &  \\ 
        &  &  &  &  &  & Pre-trained CNN &  \\ \hline
        
        \cite{fPA9} & S \& H & Own dataset, & ROI segmentation, & Score fusion & 5 & SVM, & S \\ 
        &  & LivDet 2017 & RGB image creation &  &  & CNN, &  \\ 
        &  &  &  &  &  & Pre-trained CNNs &  \\ \hline
        
        \cite{fPA2} & S & MSU-FPAD, & Minutiae detection, & Score fusion & 6 & Pre-trained CNN  & S \\ 
        &  & PBSKD & Patches creation, &  &  &  &  \\
        &  &  & Patches alignment &  &  &  &  \\ \hline
        
        \cite{fPA1} & S & LivDet 2011, & Dense-SIFT & Score fusion & 8 $\leq$ & SVM,  & S, U \\ 
         &  & LivDet 2013, &  &  &  & K-means, &  \\
         &  & LivDet 2015, &  &  &  & PCA &  \\
         &  & LivDet 2019 &  &  &  &  &  \\ \hline
        
        \cite{fPA8}, & W & MSU-FPAD v2, & Patches extraction & Score fusion & 3 & Pre-trained CNN & S \\ 
        \cite{fPA8_1} &  & LivDet 2015 &  &  &  &  &  \\
        &  & LiveDet 2017 &  &  &  &  &  \\ \hline
    \end{tabular}}
    \caption{Overview of Fingerprint PAD using anomaly detection for unknown PAs. (where the abbreviations used are Anomaly Detection (A. D.), Software/Hardware/Wrapper (S/H/W),  Supervised (S), Semi-Supervised (SS), and Unsupervised (U))\label{tab2}}
\end{table}
We now review the related work for FPAD in general, and specifically for unknown attack detection of fingerprints. Many software and hardware PAD methods are presented in the literature to detect PAs against fingerprint recognition systems. PAs can be conducted using PAIs in two fingerprint forms (e.g., overlays), and additionally using 3d printed fingers~\cite{16}. Software approaches make use of features extracted by standard sensing technologies, which can further be divided into static (e.g., sweat pores and texture of ridges and valleys) and dynamic features (e.g., skin color change over time due to pressure). Software approaches are usually cheaper to implement (as no extra hardware is needed), and less intrusive to the user~\cite{Galbally2019}. Hardware approaches introduce a new device to the sensing technology to capture more details than standard sensors (e.g., fingerprint sweat, blood pressure, or odor). Keeping in mind that hardware solutions are only used to capture data, and they usually have associated software solutions with them that distinguish between bona fide and PAs, which can either be inbuilt in the sensor or as stand-alone software. So, in theory, if two different hardware approaches as in \cite{fPA7} and \cite{EgySWIR} use Short Wave Infrared (SWIR) and Laser Speckle Contrast Imaging (LSCI) techniques respectively, they can still process each other datasets using the same software in their approaches. According to Galbally et al.~\cite{Galbally2019} hardware-based approach introduces a higher fake detection rate than a software-based approach. This survey paper considers the type of approach (i.e., hardware and software) as a comparison factor, as shown in Table~\ref{tab2}.

\subsection{Pre-processing techniques (Software-based)}
We now briefly review the pre-processing techniques in the literature attached to the PAD methods presented in Table~\ref{tab2}. These can be texture-based descriptors such as Local Binary Pattern (LBP)~\cite{LBPPaper}, Grey Level Co-occurrence Matrix (GLCM)~\cite{GLCM}, Histogram of Oriented Gradients (HOG)~\cite{HOGPaper}, Binary Statistical Image Features (BSIF)~\cite{fPA12}, Local Phase Quantization~\cite{LPQPaper}, Binary Gabor Patterns (BGP)~\cite{LBPBGP}, Dense-SIFT~\cite{fPA1} or techniques such as Image Segmentation, Region of Interest (ROI) Segmentation or Finger-print Minutae detection.
\subsection{Convolutional Neural Network (Software-based)}
We now briefly review the deep learning-based approaches; Park et al.~\cite{fPA3} presented a supervised software-based approach using a convolution neural network (CNN), which did not use the PAD of unknown PAs. However, they tested the approach on the LivDet 2015 data sets that contains four unknown PAs~\cite{LivDet2015}. The CNN network devised by them takes the full image of a fingerprint. It outputs a three-dimensional tensor that is used to determine the probability of the image being a bona fide or an attack presentation. The liveness probability is compared to an optimal threshold obtained from the training phase, where they achieved an average classification error of 1.5\% for the unknown PAs. The usage of deep learning approaches has become a trend in the last decade, which is mainly due to the freely available pre-trained networks such as VGG~\cite{VGGPaper}, GoogleNet~\cite{GoogleNetPaper}, and ResNet~\cite{ResnetPaper}. 
Tolosana et al.~\cite{fPA9} published a new experiment, where a PAD method relies on the use of SWIR and RGB images. Deep features from RGB images are extracted via two pre-trained CNNs, namely VGG19 and MobileNet, and a ResNet network trained from scratch. The features output by the CNNs is feed to an SVM. Additionally, handcrafted features as spectral signatures were extracted from SWIR images. For the final evaluation, a score fusion applied, and the reported D-EER for this experiment was 1.36\%.

\subsection{Known Sensor \& Known Attacks}
Marasco et al.~\cite{RossSurveyFPAD} provided an overview of PAD methods in the literature for fingerprint recognition systems, and they specifically point out that commercial fingerprint recognition systems can be spoofed. Most of these approaches test their performance on a test dataset with the same PAs as used during the training. Thus, these PAs are considered known to the PAD-method, which is a less realistic scenario than a real-world environment setup where additional PAIs may be used to conduct PAs (i.e., unknown attacks). 
\subsection{Known Sensor \& Unknown Attacks}
To the best of our knowledge, Tan et al.~\cite{fPA00} were the first to point to the effect of environmental conditions and new PAI materials on PAD methods for fingerprints. They showed that new PAI to increases the error rate by at least 14\% and up to 55,6 \% on different fingerprint scanners as Identix, Crossmatch, and Digital Persona. Moreover, their experiment showed that the error rate drops back into an acceptable range once new PAIs are used in the training phase. This was later confirmed by Marasco et al. in~\cite{fPA10}, in which they experimented the increase of spoof detection error rates of five fingerprint liveness detection methods (given by Marasco et al.~\cite{1of5}, Moon et al.~\cite{2of5}, Shankar et al.~\cite{3of5}, Abhyankar et al.~\cite{4of5}, and Tan et al.~\cite{5of5}) when tested on the new PAIs that were not used during training. Marasco et al.~\cite{fPA10} used the leave-one-out approach in their experiment, where only one PAI out of gelatine, play-doh, and silicone is used for testing, and the other two are used for training as they train the PAD methods using both PAs and bona fide presentations and can be classified as supervised anomaly detection approach.
To solve the problem of unknown PAIs, Rattani et al.~\cite{fPA12} proposed a scheme for automatic detection and adoption of the liveness detector to new PAIs. Their liveness detection is a combination of a multiclass-SVM and rule-based approaches that form an AdaBoost-based classifiers\cite{ada}. The Adaboost classifiers are used to detect novel PAs, and new PAIs used in each attack, followed by a binary classification SVM that corresponds to live and spoof classes, where the thresholds are maintained by multi-class SVM. In a case where a novel PA is presented to the detector, two rules apply to determine whether the PA is novel or already known. The first rule computes the maximum posterior probabilities for each known PA and bona fide. So, PA is considered novel if it overcomes a defined threshold else it is regarded as a known PAs and belongs to the corresponding class value. The second rule estimates the standard deviation of the posterior probabilities computed in the first rule. A low standard deviation value indicates doubt in classifying the PA as a known.
Additionally, they state the possibility of their PAD method to update the maintained binary classification SVM automatically, thus that it is always considered learned to known PA materials. This method is considered supervised because two out of four materials in the LiveDet 2011 dataset were used for training (i.e., two known PAIs and two unknown PAIs). The published results mentioned up to 46\% improvements in detecting unknown PAIs. Rattani et al.~\cite{fPA13} published a study where they tried to reduce the material-specific noise and apply a software-based method that learns the general artifacts in images from PAIs that correspond to different materials. This is done by using two SVMs that combine linear filtering and non-linear denoising using wavelet decomposition of PAIs on an LBP-based textural-analysis liveness detector. Their experimental results gained up to 44\% improvements in detecting unknown PAs on LiveDet 2011 dataset. The training phase during the experiment is done using one material out of five. Thus, the method is tested on four unknown attacks. Rattani et al.~\cite{fPA12} used Weibull-calibrated SVM (W-SVM) can be used both for the detection of liveness and spoofs, and discovery of new novel PAs and PAIs. Also, they claim W-SVM that supports interoperability between individual detectors. The results show 44\% improvements in detecting novel materials on Livedet 2011 dataset. Tolosona et al. ~\cite{fPA5} used a VGG pre-trained network as a PAD method in the finger recognition system. They use ShortWave Infrared Imaging (SWIR) images since the skin reflection within the SWIR spectrum of 900–1700 nm is independent of the skin tone as analyzed by the National Institute of Standards Technology (NIST). Thus, they used a hardware sensor approach to capture SWIR images of bona fide and PAs (i.e., own dataset), and a software-based approach for PAD. A total number of six unknown PAIs were detected by their PAD method, giving high convenience and secure, supervised PAD method. The same hardware developed by~\cite{fPA5} is capable of capturing finger vein images (i.e., Visible Light Images, VIS) and speckle contrast images (LSCI) in addition to SWIR images.
Gomez-Barrero et al.~\cite{fPA6} proposed a multi-modal finger PAD method where they use different ad-hoc approaches in parallel for each image type, and several SVM classifications are set to output a score of each ad-hoc approach where the final score is given by the weighted sum of all individual scores obtained. The evaluation in this approach is applied to both known and unknown PAIs (in total 35, three are unknown), resulting in a Detection Equal Error Rate (D-EER) of 2.7\%. Gomez-Barrero et al. proposed another multi-modal approach~\cite{fPA7}, in which the proposed PAD method relies on a weighted sum of two CNN networks based on SWIR images and textural and gradient information from averaged LSCI images. They applied a pre-trained VGG19 network and a ResNet network that was trained from scratch for the CNNs. The textural and gradient information extracted from averaged LSCI images is passed into three SVMs for classification. They used the dataset from~\cite{fPA6}, increasing the number of unknown attacks to five PAIs, and reporting a decrease in the D-EER from 2.7\% to 0.5\%. Chugh et al.~\cite{fPA2} proposed a software-based FPAD method with a generalization against PAIs not seen during training. They studied the characteristics of twelve different PAIs and bona fide presentations using deep features extracted by a pre-trained CNN, namely, MobileNetv1.
Further, they applied an agglomerative clustering based on the shared characteristics of PAIs. Thus, they concluded that a subset of PAIs, namely silicone, 2D paper, play-doh, gelatine, latex body paint, and monster liquid latex, are essential PAIs to include during the training to achieve a robust PAD. An android smartphone application is presented without a significant drop in performance from the original PAD method. They achieved a True Detection Rate (TDR) of 95.7\% and False Detection Rate (FDR) of 0.2 \% when the generalization set is used for training (i.e., the six PAIs).

\subsection{UnKnown Sensor \& Unknown Attacks}
Rattani et al.~\cite{fPA11} declared the need for fingerprint PAs detection to be considered as an open set recognition problem. Thus, incomplete knowledge about neither PAs nor PAIs is known to the PAD method during training. Therefore, they adopted W-SVM, which uses recent advances in extreme value theory statistics for machine learning to directly address the risk of the anomalies in an open set recognition problem. Ding et al.~\cite{Ross2016} proposed the use of an ensemble of One-Class Support Vector Machines (OC-SVM) using bona fide samples to generate a hypersphere boundary which is refined by a small number of spoof samples, for classification of unknown PAIs.  Jain et al.~\cite{JainOneClass19} developed a one-class classifier that is based on training on learning of bona fide samples using multiple GANs (Generative-Adversarial Networks) which can reject any PAI. Gonz{\'a}lez-Soler et al.~\cite{fPA1} proposed a software-based PAD method and achieved an overall accuracy by 96.17\% on the LivDet2019 competition. This method relied on three image representation approaches, which combine both local and global information of the fingerprint, namely Bag-of-words (BoW)~\cite{bag}, Fisher Vector (FV)~\cite{fisher}, and Vector Locally Aggregated Descriptors (VLAD)~\cite{vlad}. They computed Dense-SIFT descriptors at different scales, and the features are then encoded using a previously learned visual vocabulary using the previously mentioned image representation approaches. A linear SVM classifier is applied to classify the fingerprint descriptor in each method. A weighted sum computes the final decision score. BoW approach uses K-means clustering local features and presents it as a pyramid of spatial histograms. FV approach is based on statistical and spectral-based techniques, where the Gaussian Mixture Model (GMM) locates local features that lie under the same distribution. Then, these features are presented in a lower dimension via Principal Component Analysis (PCA). VLAD approach, on the other hand, relied on non-probabilistic techniques and is used to reduce the high-dimension image representation in BoW and FV. They experimented with both scenarios of their PAD method, namely supervised (i.e., known PAs) and unsupervised (i.e., unknown PAs) scenarios. Chugh et al.~\cite{fPA8} and Gajawada et al.~\cite{fPA8_1} present a wrapper that can be adopted by any fingerprint PAD method to improve the generalization performance of the PAD method against unknown PAIs. These approaches are based on synthesizing fingerprint images that correspond to unknown PAs and bona fide images as well. The goal is to transfer the characteristics into a deep feature space so that a more precise presentation helps the PAD method increase its generalization performance. The method is based on multiple pre-trained VGG19 CNNs that encode and decode the content and style loss of the synthesized images, as they can be further used to train the PAD method. They use the same PAD software method as done by Chugh et al.~\cite{fPA2} to experiment with the wrapper. Moreover, this approach is a supervised method in which they use the leave-one-out technique on each PAI for MSU-FPADv2, where the other PAIs are known in training. On the other hand, in LivDet 2017 dataset, three PAIs were considered unknown.

\section{Conclusions \& Future Directions}
\label{sec:4}
This survey paper presented unknown attack detection for fingerprints, including a survey of existing methods summarized \& categorized in Table~\ref{tab2}, additionally a taxonomy of FPAD is presented in Figure~\ref{fig:taxonomyfigure}. Currently, most unknown attack detection methods for fingerprints are solving the problem of known sensors and unknown PAIs, and there are only a few methods which are unknown sensor, and unknown PAI, including cross-dataset. \\
Unknown attack detection with unknown sensors is a relatively new area of research for FPAD and should be the focus area in near-future. The first approach to solving it is of synthesis, as done by Jain et al.~\cite{JainOneClass19}. The second approach is to arrive at a common deep-feature representation, such as the one used by Gonz{\'a}lez-Soler et al.~\cite{fPA1}. The challenge in synthesis based approach is to do high-quality synthesis of the bona fide samples, and the difficulty in arriving at a common deep-feature representation is the degree of invariance it can provide to sensor type and PAI.
\balance
{\bibliographystyle{splncs04}
\bibliography{lniguide_en}}

\end{document}